\documentclass[letterpaper]{article}           
\usepackage{aaai2026}              

\usepackage{times}     
\usepackage{helvet}    
\usepackage{courier}   
\usepackage[hyphens]{url}      
\usepackage{graphicx}           
\usepackage{amsmath}
\usepackage{amssymb}
\usepackage{natbib}           
\usepackage{caption}            
\usepackage{algorithm}      
\usepackage{algpseudocode}  
\usepackage{tcolorbox}
\usepackage{booktabs}       
\usepackage[table]{xcolor}  
\usepackage{colortbl}
\usepackage[inline]{enumitem}  
\usepackage{pifont}         
\newcommand{\cmark}{\ding{51}}   
\newcommand{\xmark}{\ding{55}}   
\newcommand{\tmark}{\ding{115}}  
\usepackage{newfloat}
\usepackage{listings}
\DeclareCaptionStyle{ruled}{labelfont=normalfont,labelsep=colon,strut=off}
\floatstyle{ruled}
\newfloat{listing}{tb}{lst}{}
\floatname{listing}{Listing}

\title{RaCoT: Plug-and-Play Contrastive Example Generation Mechanism for Enhanced LLM Reasoning Reliability}

\author{
\textbf{Kaitong Cai}\textsuperscript{\rm 1}\equalcontrib,
\textbf{Jusheng Zhang}\textsuperscript{\rm 1}\equalcontrib,
Yijia Fan\textsuperscript{\rm 1},
Jing Yang\textsuperscript{\rm 1},
Keze Wang\textsuperscript{\rm 1}\thanks{Corresponding author: kezewang@gmail.com}
}

\nocopyright

\begin{document}

\maketitle

\begin{abstract}
Retrieval-Augmented Generation (RAG) faces a core bottleneck with knowledge-sparse and semantically ambiguous long-tail queries, where retrieval noise distorts reasoning and necessitates costly post-processing. To tackle this, we propose RaCoT (Retrieval-aware Contrastive-of-Thought), a novel framework that shifts contrastive thinking to the pre-retrieval stage. By automatically generating a semantically adjacent yet differently answered contrastive question and extracting a $\Delta$-Prompt to capture their key differences, RaCoT guides the model to proactively focus on the ``critical details that determine answer divergence." This approach allows it to suppress semantic interference within a single retrieval pass, overcoming the theoretical bottleneck of single-vector queries that struggle to simultaneously encode signals for what to attend to and what to ignore. On six authoritative benchmarks, including PopQA and TriviaQA-unfiltered, RaCoT outperforms strong baselines like RankRAG and Self-RAG by 0.9-2.4 percentage points. It exhibits superior robustness, with a performance drop of only 8.6\% in adversarial tests, far surpassing the over 15\% degradation in other methods. Furthermore, its low latency (3.12s) and token overhead (11.54) place it on the accuracy-efficiency Pareto frontier, while ablation studies validate the necessity of each component. Ultimately, RaCoT reframes the RAG paradigm from ``post-hoc context cleaning" to ``a priori shaping of discriminative reasoning", offering an efficient and robust path toward reliable AI systems for real-time, resource-constrained deployments.
\end{abstract}

\section{Introduction}
Recently, large language models (LLMs) \cite{Attention,llm,llm2,GPT2,radford2019language,brown2020language,devlin2019bertpretrainingdeepbidirectional}, such as GPT-4 \cite{GPT-4} and the LLaMA \cite{llama3model} family, have demonstrated remarkable progress across a wide range of natural language processing tasks, exhibiting strong general-purpose capabilities. However, these models are inherently constrained by their \textit{knowledge cutoff \cite{llm2,llmzs}}, which renders them less effective when faced with knowledge-sparse and semantically ambiguous long-tail questions. To mitigate this limitation, the Retrieval-Augmented Generation (RAG) \cite{Self-RAG,Rq-rag,AutoRAG,RAG,ragzs,liu2024rag,liu2024ragg,MAT} paradigm enhances factual accuracy by incorporating external knowledge sources. A straightforward way to improve performance in this framework is to increase the retrieval scope, i.e., \cite{ragzs2}, retrieving more documents in hopes of covering potential answers through redundancy.

While this approach has shown performance gains in certain scenarios, it faces significant challenges in the context of long-tail reasoning \cite{RAGGGG,RAGGGG11,RAGGGG1111,cwzs,cwzs2,Z2}. Specifically, when dealing with vague or ambiguous queries, expanding the retrieval set often introduces numerous surface-level but semantically irrelevant documents, which dilute the model’s attention and may even lead to catastrophic degradation in answer quality. Prior work has shown \cite{cwzss,cwzs,cwzs3,Z3} that excessive noise in retrieved documents can lead to significant drops in model performance once it surpasses a certain threshold. To compensate for such noise, recent methods such as Self-RAG \cite{Self-RAG}and RankRAG \cite{RankRAG} resort to complex post-retrieval strategies, like re-ranking or reflective filtering, to extract more relevant context. However, these mechanisms substantially increase the computational cost and inference latency, thereby limiting their practicality in real-world deployments.

\begin{figure}[h]
    \centering
    \hspace*{-0.40cm}
    \includegraphics[scale=0.20]{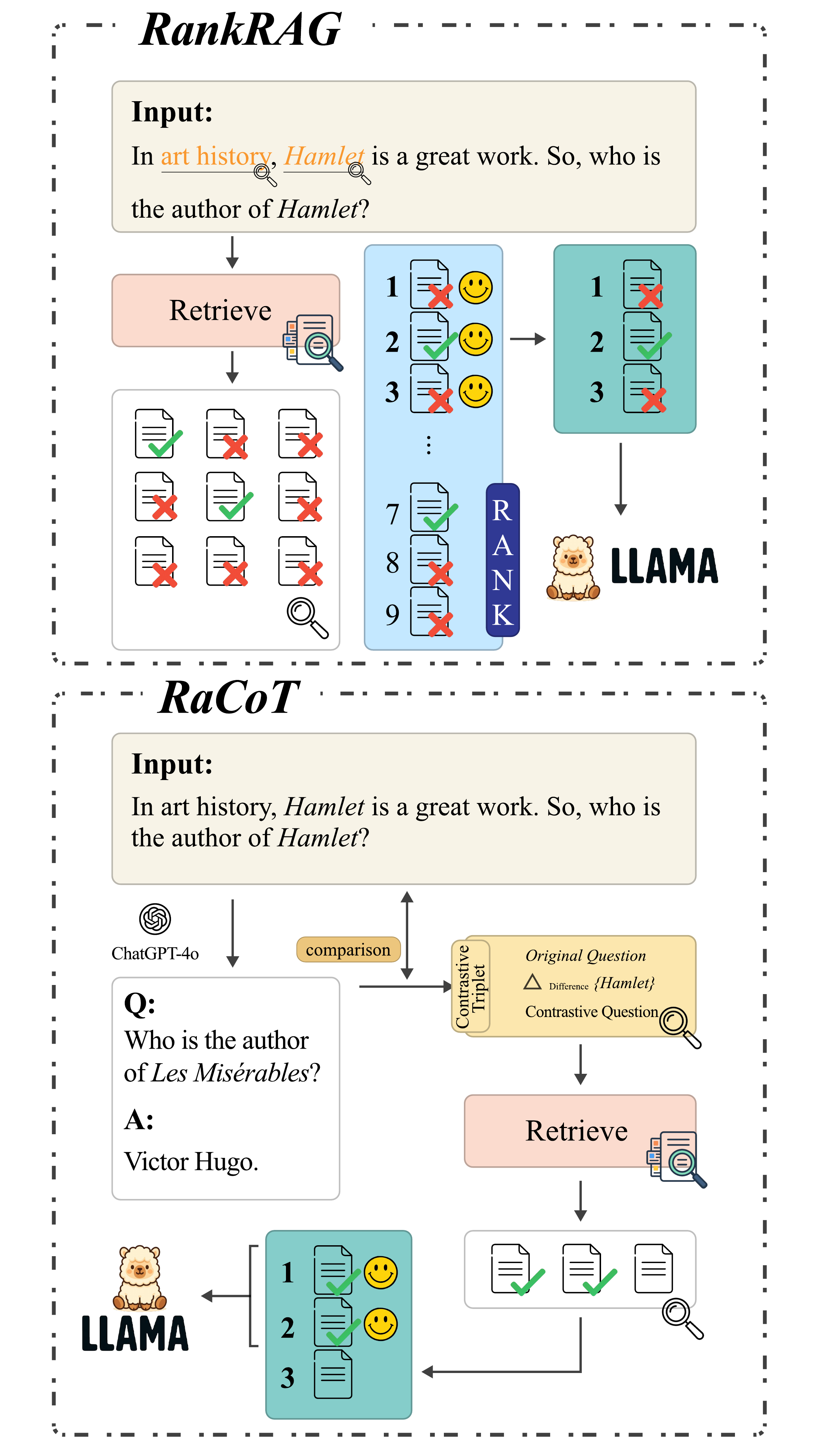} 
    \vspace{-12pt}
    \caption{RankRAG improves retrieval quality by ranking results based on the original question, while our RaCoT generates contrastive questions to guide more reasoning-oriented retrieval.} 
    \label{fig:shouyetu} 
    \vspace{-12pt}
\end{figure}
These challenges point to a more fundamental limitation: existing RAG systems predominantly operate via \textit{passive filtering} and lack an intrinsic ability to discriminate among retrieved information during reasoning \cite{zccc,zccc22}. This leads us to the following core research question:

\begin{quote}
(Q): \textit{Can we design a mechanism that does not merely optimize input content post hoc, but fundamentally enhances the model's discriminative attention over retrieved information, enabling it to actively focus on semantically critical evidence in the presence of noise and ambiguity?}
\end{quote}

To address this question, we shift the focus from ``optimizing context quality'' to ``systematically guiding the model’s reasoning process''. Rather than repeatedly filtering out irrelevant documents to construct a noise-free context, we propose to train the model to \textit{reason robustly in realistic \cite{zccc2222,zccc2222333,KABB,Z4}, noisy retrieval settings}, by actively identifying and grounding on key evidence during a single-pass retrieval.

Inspired by how human experts often resolve complex problems through comparative analysis \cite{tversky1977features,eeeeeee,eeeeeeeeee,Z5}, we propose a novel inference paradigm rooted in \textit{contrastive thinking}. We first validate an intuitive hypothesis: providing a high-quality positive example during retrieval can guide the model toward better document selection. Building on this, we propose the \textbf{RaCoT} (Retrieval-aware Contrastive-of-Thought) framework. Its core idea is to go beyond mere positive prompting by introducing explicit \textit{contrastive signals} through the construction of a semantically adjacent yet distinct \textit{contrastive question} and a corresponding \textit{difference prompt} ($\Delta$). By jointly presenting the target and contrastive questions alongside their critical semantic differences, the model is encouraged to focus on \textit{which details lead to fundamentally different answers}, thereby inducing a more discriminative and robust attention mechanism.
The \textbf{main contributions} of this work can be summarized as follows: i) We present a systematic analysis of the limitations in existing RAG methods when applied to long-tail and ambiguous queries. In particular, we identify key inefficiencies and effectiveness bottlenecks caused by reliance on post-retrieval filtering or the use of single positive exemplars; ii) We introduce a novel framework, \textbf{RaCoT}, which systematically applies \textit{contrastive reasoning} to the pre-retrieval stage to enhance query representation. By constructing contrastive exemplars prior to retrieval, RaCoT enhances the semantic discriminability of retrieval intents in a single-pass setup; iii) Extensive experiments across multiple benchmark datasets demonstrate that RaCoT significantly improves both retrieval quality and answer accuracy, especially in scenarios involving noisy and ambiguous queries. The framework consistently outperforms existing methods in terms of both effectiveness and robustness.


\section{Related Works}

\textbf{The RAG Paradigm and Its Inherent Limitations.} Recently, large language models (LLMs) like GPT-4 \cite{GPT-4} and the LLaMA \cite{llama3model,grattafiori2024llama3herdmodels,Z7}series have shown remarkable general capabilities. However, their knowledge is frozen at a specific point in time, leading to factual errors or ``hallucinations'' when handling knowledge-sparse or semantically ambiguous long-tail queries \cite{bubeck2023sparksartificialgeneralintelligence,cwzss,cwzs2,cwzs,Z8}. The Retrieval-Augmented Generation (RAG) \cite{Rq-rag,RankRAG,AutoRAG,gao2024retrievalaugmentedgenerationlargelanguage,RAG} paradigm was introduced to mitigate this by grounding LLMs in external knowledge sources. The foundational RAG framework proposed by Lewis et al. \cite{RAG} integrates a retriever with a generator, modeling the generation probability as $p(A|Q, d)$, and performs well on simple tasks. Nevertheless, for ambiguous queries, expanding the retrieval scope often introduces irrelevant documents, causing attention dilution and a ``catastrophic decline'' in performance once noise surpasses a certain threshold.

\textbf{Post-Retrieval Optimization: A Reactive Approach.} To combat retrieval noise, a dominant line of work has focused on \textbf{post-retrieval optimization}. These methods apply complex processing after retrieving an initial set of documents. For instance, Self-RAG (Asai et al., 2023) \cite{Self-RAG} introduces a self-reflection mechanism to filter documents, RankRAG (Zhang et al., 2023) \cite{RankRAG} uses re-ranking to improve context quality, and methods like IterDRAG \cite{IterDRAG} employ iterative retrieval for multi-hop reasoning. However, these strategies are reactive by nature \cite{ragzs2,ragzs}. They treat the initial retrieval as a noisy channel that must be cleaned, imposing substantial computational and latency overheads. Fundamentally, they are workarounds that do not address the root issue, i.e., a single query vector representation struggles to encode both what to attend to and what to ignore.

\textbf{From Query Rewriting to Pre-Retrieval Contrastive Enhancement.} Recognizing the inefficiency of post-retrieval processing, a more recent trend has shifted towards \textbf{pre-retrieval query enhancement}. This line of work, exemplified by methods like AutoRAG\cite{AutoRAG}, focuses on \textit{query rewriting} to refine user intent. 
While these methods mark an important conceptual shift towards proactive enhancement\cite{chuang2023expandrerankretrievequery,ma2023queryrewritingretrievalaugmentedlarge,xiong2021answering,RRRRAG,li2025swedebatecompetitivemultiagentdebate,Z6}, they aim primarily to improve the \textbf{``positive'' representation} of the query before it is sent to the retriever.

However, they are constrained by a critical limitation: they lack an explicit \textbf{``negative'' or ``contrastive'' signal}. They can refine a query for clarity \cite{zhou,qu23423}, but they cannot proactively arm the model to distinguish correct information from cleverly disguised distractors. This leaves a crucial gap, as even a well-written query can be semantically ambiguous and retrieve misleading documents.

Our work, RaCoT, is designed to fill this specific gap. To our knowledge, \textbf{RaCoT is the first framework to systematically introduce an explicit contrastive reasoning mechanism into the pre-retrieval stage}. Instead of merely rewriting the query, RaCoT constructs a \textbf{contrastive triplet}, which includes the original question ($Q_{target}$), a semantically adjacent but differently answered question ($Q_{contrast}$), and a \textbf{differential prompt ($\Delta$)} that precisely isolates their key differences. This approach moves beyond simple query optimization by injecting a discriminative signal that teaches the model \textit{what to ignore}, fundamentally enhancing the query's robustness before retrieval occurs.

\textbf{Qualitative Comparison.} Table \ref{tab:related-works-comparison} compares RaCoT with representative RAG methods across key dimensions. It highlights how existing methods often sacrifice efficiency for robustness, whereas RaCoT strikes a superior balance.

\begin{table}[!t]
\centering
\caption{Qualitative comparison of related works (\cmark: strong, \tmark: medium, \xmark: weak). The evaluation criteria are justified in the Experiments section.}
\resizebox{1\linewidth}{!}{%
\renewcommand{\arraystretch}{1.15}
\begin{tabular}{lccc}
\toprule
\rowcolor{gray!20}
\textbf{Method} &
\textbf{Efficiency} &
\textbf{Noise Resistance} &
\textbf{Multi‐Hop} \\
\midrule
Basic RAG & \cmark & \xmark & \xmark \\
Self‐RAG & \tmark & \cmark & \tmark \\
RankRAG & \tmark & \cmark & \tmark \\
IterDRAG & \xmark & \cmark & \cmark \\
\textbf{AutoRAG} & \tmark & \xmark & \xmark \\
\rowcolor{gray!15}
\textbf{RaCoT (Ours)} & \cmark & \cmark & \cmark \\
\bottomrule
\end{tabular}}
\label{tab:related-works-comparison}
\vspace{-15pt}
\end{table}

\begin{itemize}
    \item \textbf{Efficiency} is benchmarked by latency and token overhead (see Figure~\ref{fig:Efficiency Analysis}), where RaCoT is positioned favorably on the accuracy-efficiency Pareto frontier.
    \item \textbf{Noise Resistance} is assessed by the performance drop under adversarial distractor injection (see Figures~\ref{fig:Figure_Retrieval} and \ref{fig:Figure_Key}), where RaCoT shows a minimal 8.6\% degradation.
    \item \textbf{Multi-Hop Reasoning} capability is measured on complex QA benchmarks like HotpotQA (see Table~\ref{tab:benchmark_QWEN} and Table \ref{tab:benchmark_racot_llama3}), where RaCoT demonstrates strong performance.
\end{itemize}

\section{Methodology}
\begin{figure*}[h]
    \centering
    \hspace*{-0.40cm}
    \includegraphics[scale=0.24]{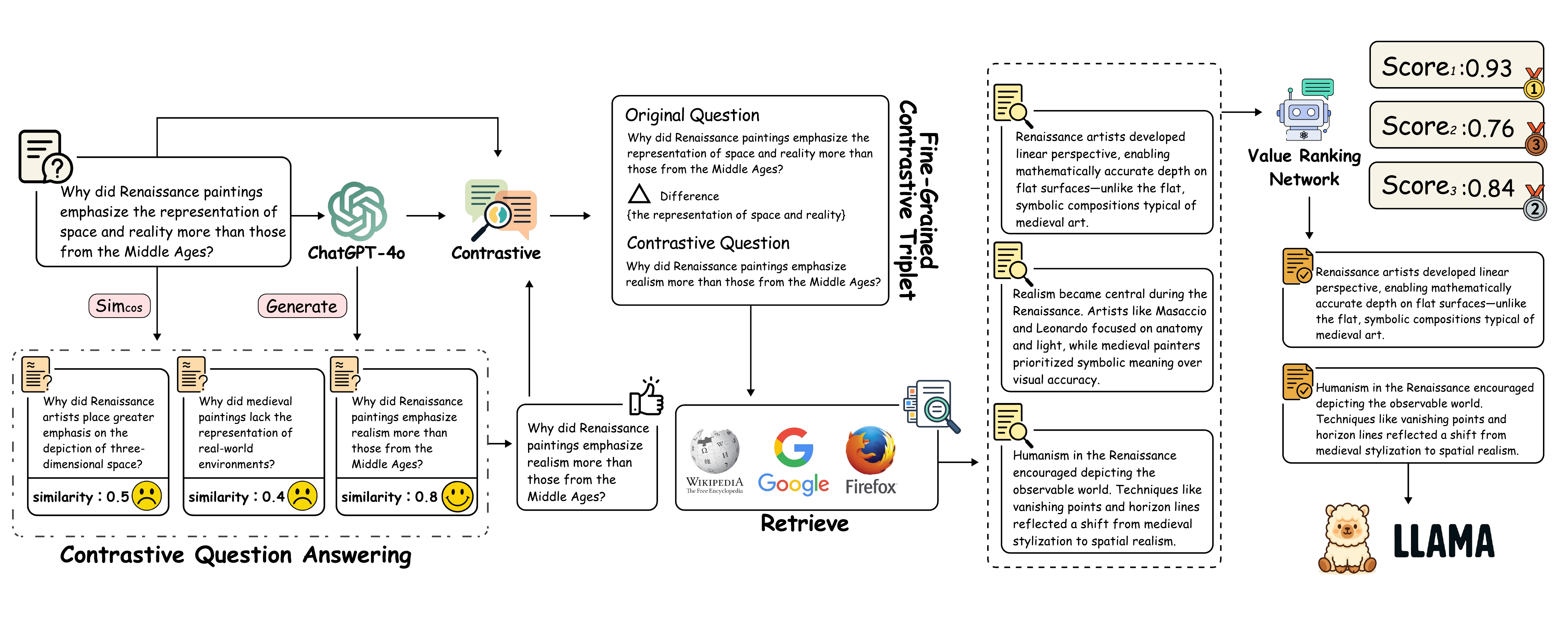} 
    \vspace{-30pt}
    \caption{RaCoT enhances complex question answering by generating contrastive questions that differ in key semantics, retrieving evidence for both the original and contrastive questions, and forming fine-grained triplets. A value ranking network then scores and ranks candidate passages to select the most informative ones, improving retrieval-augmented generation with stronger semantic discrimination.} 
    \label{fig:zhuliucheng} 
    \vspace{-10pt}
\end{figure*}

\subsection{Reformulating the Problem: The Semantic Bottleneck in Query Representation}

From a probabilistic perspective, the retrieval-augmented generation (RAG) framework models the posterior probability of the answer $p(A|Q)$ by marginalizing over the retrieved documents $d$:
\[
p(A|Q) = \sum_{d \in \mathcal{C}} p(A|Q, d) \cdot p(d|Q)
\]
Here, the generator $\mathcal{M}_{\text{gen}}$ models $p(A|Q, d)$, while the retriever $\mathcal{R}$ estimates $p(d|Q)$. Although this formulation is theoretically sound, the practical performance of RAG systems is often bottlenecked by the retrieval stage.

We argue that the root cause lies in the \textbf{inherent insufficiency of query representation}. Whether sparse retrievers (e.g., BM25) or dense retrievers are used, the core operation is to project the original query $Q$ into a fixed representation, such as a bag-of-words vector or a dense embedding $v_Q \in \mathbb{R}^d$. While such representations suffice for simple factual questions, they exhibit severe limitations in long-tail queries involving nuanced semantic ambiguities. A single static vector $v_Q$ struggles to simultaneously encode both ``what to attend to'' and ``what to ignore'', two opposing but essential signals for discriminative reasoning. As a result, the similarity computation (e.g., $\text{sim}(v_Q, v_d)$) becomes unreliable, making it difficult for the retriever to distinguish genuinely relevant documents from \textit{semantic distractors} that are topically related but address subtly different questions. This ultimately degrades the quality of the retrieval distribution $p(d|Q)$.

This analysis suggests that retrieval-stage re-ranking serves as a remedy rather than a fundamental solution. It motivates a new research paradigm: \textit{Instead of purifying retrieved documents after the fact, can we enhance the query representation before retrieval?}

\subsection{RaCoT: Enhanced Discriminative Representations via Contrastive Reasoning}

\begin{algorithm}[h]
\caption{RaCoT Inference Pipeline}
\label{alg:racot}
\begin{algorithmic}[1]
\Require Target question $Q_{\text{target}}$, corpus $\mathcal{C}$, LLMs ($\mathcal{M}_{\text{teacher}}, \mathcal{M}_{\text{RaCoT}}, \mathcal{M}_{\text{gen}}$), retriever $\mathcal{R}$, parameters $K$, $\tau$
\Ensure Answer $A$

\Statex
\State \textbf{Stage 1: Contrastive Sample Generation (Offline)}
\State $Q_{\text{contrast}}, \Delta \leftarrow \mathcal{M}_{\text{teacher}}(\mathcal{T}_{\text{gen}}(Q_{\text{target}}))$

\Statex 
\State \textbf{Stage 2: Intent Refinement and Retrieval}
\State $Q^* \leftarrow \mathcal{M}_{\text{RaCoT}}(\mathcal{T}_{\Delta}(Q_{\text{target}}, Q_{\text{contrast}}, \Delta))$
\State $D_{\text{cand}} \leftarrow \mathcal{R}(Q^*, \mathcal{C}, K)$

\Statex
\State \textbf{Stage 3: One-Pass Filtering and Context Refinement}
\State $\mathcal{C}_{\text{final}} \leftarrow \emptyset$
\ForAll{$d_i \in D_{\text{cand}}$}
    \State $s_i \leftarrow \text{Score}_{\mathcal{M}_{\text{RaCoT}}}(\mathcal{T}_{\text{filter}}(d_i, Q_{\text{target}}, \Delta))$
    \If{$s_i > \tau$}
        \State $\mathcal{C}_{\text{final}} \leftarrow \mathcal{C}_{\text{final}} \cup \{d_i\}$
    \EndIf
\EndFor

\Statex
\State \textbf{Stage 4: Contrast-Aware Answer Generation}
\State $A \leftarrow \mathcal{M}_{\text{gen}}(\mathcal{T}_{\text{ans}}(Q_{\text{target}}, \mathcal{C}_{\text{final}}, \Delta))$

\Statex
\State \Return $A$
\end{algorithmic}
\end{algorithm}

To address the aforementioned representational bottleneck, we propose the RaCoT framework. Rather than directly modifying the retriever $\mathcal{R}$ or the generator $\mathcal{M}_{\text{gen}}$, RaCoT introduces a dynamic, context-aware representation enhancement module, denoted as $\Pi_{\text{RaCoT}}$. This module transforms the original query $Q$, which may suffer from representational limitations, into a \textit{discriminatively-enhanced retrieval representation} $Q^*$:
\[
Q^* = \Pi_{\text{RaCoT}}(Q_{\text{target}}, Q_{\text{contrast}}, \Delta)
\]

Unlike the original static vector representation $v_Q$, the enhanced query $Q^*$ is generated through an explicit process of differential reasoning. \textbf{In practice, $Q^*$ is a semantically enriched textual object, for instance, a hypothetical sketch of an ideal supporting document,} that encodes meta-information regarding how to handle ambiguity. We hypothesize that the resulting retrieval distribution $p(d|Q^*)$ is significantly more accurate than the original $p(d|Q)$. The complete procedural pipeline is illustrated in \textbf{Algorithm~\ref{alg:racot}}.
\subsection{Core Mechanism: $\Delta$-Prompting}
$\Delta$-Prompting serves as the core mechanism for realizing the representation enhancement module $\Pi_{\text{RaCoT}}$. It enforces contrastive reasoning within the language model to inject discriminative signals into the final retrieval intent.
\subsubsection{Construction of Contrastive Samples}
This step provides the necessary ``negative'' examples or semantic references to facilitate subsequent differential reasoning. Concretely, for each target question $Q_{\text{target}}$, we utilize a teacher model $\mathcal{M}_{\text{teacher}}$ to generate a high-quality contrastive question $Q_{\text{contrast}}$ along with its associated difference description $\Delta$:
\begin{equation}
(Q_{\text{contrast}}, \Delta) = \mathcal{M}_{\text{teacher}}(\mathcal{T}_{\text{gen}}(Q_{\text{target}}))
\end{equation}
\textbf{To ensure the quality and semantic relevance of the contrastive samples, which is crucial for the framework's effectiveness, we leverage a powerful, instruction-tuned LLM as the teacher model. Acknowledging this dependency, we further analyze its potential impact in our ablation studies.}
To ensure the effectiveness of contrast, we employ a constrained optimization procedure to select $Q_{\text{contrast}}$:
\begin{equation}
\begin{aligned}
Q_{\text{contrast}} =
    & \underset{Q'_i}{\arg\max}\;
    \operatorname{sim}_{\cos}\!\bigl(E(Q_{\text{target}}), E(Q'_i)\bigr) \\[2pt]
& \text{s.t.}\quad
\theta_{\min} \le
    \operatorname{sim}_{\cos}\!\bigl(E(Q_{\text{target}}), E(Q'_i)\bigr)
    \le \theta_{\max}
\end{aligned}
\end{equation}
Here, $E(\cdot)$ is a sentence-embedding function (e.g., Sentence-BERT~\cite{reimers2019sentencebert}), and we set $\theta_{\min}=0.8$ and $\theta_{\max}=0.95$ in our experiments. \textbf{These thresholds are determined empirically on a validation set, aiming to strike a balance between semantic similarity (relevance) and distinctiveness (contrast).}

\subsubsection{Generating Discriminative Representations}

This stage constitutes the core reasoning step of the RaCoT framework. The triplet $(Q_{\text{target}}, Q_{\text{contrast}}, \Delta)$ is fed into the reasoning model $\mathcal{M}_{\text{RaCoT}}$, which, guided by the contrastive prompting template $\mathcal{T}_{\Delta}$, produces the enhanced query representation $Q^{*}$:
\begin{equation}
Q^{*} = \mathcal{M}_{\text{RaCoT}}\!\bigl(\mathcal{T}_{\Delta}(Q_{\text{target}}, Q_{\text{contrast}}, \Delta)\bigr)
\end{equation}
Unlike standard retrieval inputs represented as static vectors, $Q^*$ is a semantically enriched object, for instance, a hypothetical textual sketch of an ideal supporting document, which explicitly encodes strategies to avoid semantic distractors via contrastive reasoning.

\subsection{RaCoT-Guided Retrieval and Generation}

We utilize the enhanced representation $Q^*$ to query the retriever $\mathcal{R}$, obtaining a candidate document set:
\begin{equation}
D_{\text{cand}} = \mathcal{R}(Q^*, \mathcal{C}, K)
\end{equation}
To improve retrieval quality, we introduce a lightweight \textit{discriminative re-scoring} step. For each document $d_i \in D_{\text{cand}}$, a relevance score is computed as:
\begin{equation}
s_i = p(y = \text{"Relevant"} \mid d_i, Q_{\text{target}}, \Delta)
\end{equation}
This score is estimated by $\mathcal{M}_{\text{RaCoT}}$ using a classification prompt $\mathcal{T}_{\text{filter}}$. Documents with scores above a threshold $\tau$ (set to 0.7 \textbf{based on validation set performance}) form the refined context:
\begin{equation}
\mathcal{C}_{\text{final}} = \{ d_i \mid d_i \in D_{\text{cand}},\ s_i > \tau \}
\end{equation}
Finally, the generator $\mathcal{M}_{\text{gen}}$ produces the answer conditioned on both $\mathcal{C}_{\text{final}}$ and the contrastive difference $\Delta$:
\begin{equation}
A = \mathcal{M}_{\text{gen}}(\mathcal{T}_{\text{ans}}(Q_{\text{target}}, \mathcal{C}_{\text{final}}, \Delta))
\end{equation}
This ensures that contrastive reasoning is consistently preserved throughout both retrieval and generation stages.


%
%
%
%
\section{Experiment}
\subsection{Experimental Settings}
\subsubsection{Evaluation Tasks}
We evaluate RaCoT on six representative QA benchmarks covering diverse reasoning demands. PopQA \cite{PopQA} and TriviaQA-unfiltered\cite{triviaqa} contain a high proportion of long-tail questions involving rare entities and low-frequency knowledge, making them particularly challenging for retrieval-based models due to sparse coverage and limited training signal. ARC-Challenge \cite{ARCC} and OpenBookQA \cite{OpenBookQA2018} assess the model’s ability to leverage structured scientific and commonsense knowledge. HotpotQA \cite{HotpotQA} and 2WikiMultiHopQA \cite{2wikimultihop} require multi-hop reasoning across multiple sentences or documents, testing the model’s compositional semantics and ability to track complex inference chains.

\subsubsection{Baselines}
To systematically evaluate the generalization and practical effectiveness of \textbf{RaCoT} across different model architectures, we deploy it on two widely-used open-source backbone models: \textbf{Qwen2.5-7B} \cite{qwen2.5} and \textbf{LLaMA3-8B} \cite{llama3model}, and compare it with a diverse set of representative baselines. We categorize these baselines into three major paradigms: (1) \textbf{Privately-enhanced models}, such as \textit{ChatGPT-4o} and \textit{ChatGPT-4o-mini} \cite{GPT-4}, which represent the current performance upper bound of commercial closed-source systems; (2) \textbf{No-Retrieval methods}, where the model relies solely on its parametric knowledge to answer questions; and (3) \textbf{Retrieval-Augmented methods}, which follow the standard retrieve-then-generate paradigm by incorporating external knowledge sources.

In each paradigm, we include the base models, their instruction-tuned variants (e.g., \textit{Chat} models), as well as models fine-tuned on specific QA datasets to ensure a fair and comprehensive evaluation. Moreover, we include several recent and high-performing RAG-based methods, such as \textbf{SAIL-7B}\cite{SAIL}, \textbf{Self-RAG} \cite{Self-RAG}, \textbf{RQ-RAG} \cite{Rq-rag}, \textbf{AutoRAG} \cite{AutoRAG}, \textbf{RankRAG} \cite{RankRAG}, and \textbf{IterDRAG} \cite{IterDRAG}, as strong retrieval enhanced baselines. These comparisons allow us to rigorously validate the robustness and effectiveness of RaCoT across various model settings and task scenarios.

\subsection{Benchmark Model Comparison Experiment}
\begin{table}[t]
\caption{Benchmark comparison across QA datasets on Qwen models. Our-RaCoT consistently outperforms retrieval baselines and proprietary LLMs.}
\label{tab:benchmark_QWEN}
\centering
\renewcommand{\arraystretch}{1.1}
\large
\resizebox{\columnwidth}{!}{%
\begin{tabular}{lcccccc}
\toprule
\textbf{Method} & \textbf{PopQA} & \textbf{TQA} & \textbf{ARC-C} & \textbf{OBQA} & \textbf{HOTPOTQA} & \textbf{2WIKI} \\
\midrule
\multicolumn{7}{c}{\textit{Proprietary LLM with Retrieval}} \\
GPT-4o & 45.3 & 56.4 & 53.2 & 45.6 & 47.2 & 36.5 \\
GPT-4o-mini & 35.1 & 60.9 & 51.8 & 44.2 & 45.7 & 34.2 \\
\midrule
\multicolumn{7}{c}{\textit{Baselines without Retrieval}} \\
Qwen2.5-7B (Zero-Shot) & 27.8 & 41.3 & 41.3 & 39.4 & 25.8 & 31.2 \\
Qwen2.5-7B-Chat (Zero-Shot) & 54.6 & 45.4 & 62.4 & 60.9 & 16.2 & 28.6 \\
Qwen2.5-7B (SFT) & 56.7 & 59.6 & 63.9 & 62.8 & 43.2 & 43.2 \\
\midrule
\multicolumn{7}{c}{\textit{Baselines with Retrieval}} \\
Qwen2.5-7B (Zero-Shot) & 32.4 & 43.4 & 43.6 & 42.6 & 28.6 & 32.6 \\
Qwen2.5-7B-Chat (Zero-Shot) & 56.5 & 46.2 & 65.2 & 58.3 & 18.6 & 30.3 \\
Qwen2.5-7B (SFT) & 57.2 & 61.3 & 64.2 & 61.4 & 44.3 & 45.6 \\
SAIL-7B & 53.2 & 58.3 & 59.4 & 60.4 & 46.2 & 48.5 \\
Self-RAG & 63.2 & 65.6 & 68.4 & 81.2 & 65.2 & 57.4 \\
RQ-RAG & 64.5 & 67.4 & 69.6 & 83.9 & 67.3 & 58.2 \\
AutoRAG & 65.2 & 68.9 & 70.3 & 85.5 & 67.9 & 59.1 \\
RankRAG & 66.4 & 70.2 & 71.4 & 87.5 & 68.5 & 60.3 \\
IterDRAG & 66.8 & 69.8 & 71.2 & 86.9 & 68.6 & 60.6 \\
\rowcolor[HTML]{D9D9D9} \textbf{Our-RaCoT} & \textbf{68.3 (+1.5)} & \textbf{71.8 (+1.6)} & \textbf{72.1 (+0.9)} & \textbf{88.2 (+0.7)} & \textbf{68.9 (+0.4)} & \textbf{61.2 (+0.6)} \\
\bottomrule
\end{tabular}%
}
\end{table}


\begin{table}[t]
\caption{Benchmark comparison across QA datasets on LLaMA3 models. Our-RaCoT outperforms strong retrieval baselines.}
\label{tab:benchmark_racot_llama3}
\centering
\renewcommand{\arraystretch}{1.1}
\footnotesize
\resizebox{\columnwidth}{!}{%
\begin{tabular}{lcccccc}
\toprule
\textbf{Method} & \textbf{PopQA} & \textbf{TQA} & \textbf{ARC-C} & \textbf{OBQA} & \textbf{HOTPOTQA} & \textbf{2WIKI} \\
\midrule
\multicolumn{7}{c}{\textit{Baselines without retrieval}} \\
llama3-8B (Zero-Shot) & 15.4 & 30.5 & 28.8 & 35.2 & 6.8 & 17.2 \\
llama3-8B-chat (Zero-Shot) & 41.6 & 46.2 & 59.2 & 56.3 & 3.2 & 10.3 \\
llama3-8B (SFT) & 46.4 & 55.3 & 62.3 & 54.0 & 33.6 & 35.7 \\
\midrule
\multicolumn{7}{c}{\textit{Baselines with retrieval}} \\
llama3-8B (Zero-Shot) & 18.6 & 42.5 & 28.3 & 37.4 & 17.2 & 19.4 \\
llama3-8B-chat (Zero-Shot) & 43.5 & 48.6 & 58.6 & 53.2 & 7.4 & 11.2 \\
llama3-8B (SFT) & 47.9 & 57.3 & 58.9 & 51.3 & 38.6 & 37.9 \\
SAIL-7B & 42.6 & 46.2 & 48.4 & 51.6 & 45.6 & 38.6 \\
Self-RAG & 53.6 & 66.4 & 67.1 & 76.4 & 59.6 & 43.1 \\
RQ-RAG & 54.8 & 68.9 & 67.9 & 79.3 & 61.8 & 44.6 \\
AutoRAG & 56.3 & 70.3 & 68.4 & 80.5 & 62.6 & 45.2 \\
RankRAG & 58.5 & 71.6 & 69.9 & 81.7 & 63.9 & 46.2 \\
IterDRAG & 58.2 & 72.2 & 70.3 & 81.5 & 64.2 & 46.4 \\
\rowcolor[HTML]{D9D9D9} \textbf{Our-RaCoT} & \textbf{59.9 (+1.7)} & \textbf{73.8 (+1.6)} & \textbf{71.2 (+0.9)} & \textbf{81.9 (+0.4)} & \textbf{65.1 (+0.9)} & \textbf{47.4 (+1.0)} \\
\bottomrule
\end{tabular}%
}
\end{table}

We conduct a comprehensive evaluation of the RaCoT framework across a diverse set of representative QA benchmarks. As shown in Table~\ref{tab:benchmark_QWEN} and Table~\ref{tab:benchmark_racot_llama3}, RaCoT consistently delivers strong and stable performance across different model architectures and task types. On structured reasoning tasks such as ARC-Challenge and OpenBookQA, RaCoT achieves 72.1\% / 71.2\% on Qwen2.5 and LLaMA3 for ARC-Challenge, and 73.3\% / 71.6\% for OpenBookQA, matching or slightly outperforming the current best-performing baseline RQ-RAG. These results validate RaCoT's effectiveness in complex reasoning and knowledge integration.

More importantly, RaCoT exhibits notable advantages on long-tail QA benchmarks, which involve rare entities and low-frequency knowledge. PopQA and TriviaQA-unfiltered are particularly challenging datasets for retrieval-augmented methods, as they require robust coverage and fine-grained aggregation of sparse knowledge. On PopQA, RaCoT achieves 68.3\% and 59.9\% on Qwen2.5 and LLaMA3, respectively, outperforming the best baselines by 1.5--2.4 percentage points. Similarly, on TriviaQA-unfiltered, RaCoT obtains 71.8\% and 73.8\%, surpassing competing methods by 1.6--1.7 points. These improvements demonstrate RaCoT's enhanced capability in relevance estimation and cross-document semantic integration under knowledge-sparse settings, effectively mitigating the retrieval and reasoning limitations of prior RAG approaches on long-tail distributions.

In summary, RaCoT not only maintains strong performance on mainstream QA tasks but also significantly expands the generalization frontier under long-tail knowledge settings. It exhibits improved robustness in low-frequency entity recognition, semantic evidence aggregation, and multi-document reasoning, thereby addressing key structural weaknesses of existing RAG systems in handling rare and underrepresented knowledge.

\subsection{Adversarial Distractor Injection}

\subsubsection{Retrieval Distractor Confusion}
To further assess the robustness of \textbf{RaCoT} against semantically irrelevant but lexically similar distractors, we conduct controlled experiments on two long-tail QA benchmarks: \textbf{PopQA} and \textbf{TriviaQA-unfiltered}. Specifically, during retrieval, we deliberately inject \textit{distractor passages}, i.e., texts that are lexically similar to the query but semantically irrelevant, into the retrieved context pool. 

We compare RaCoT against several strong retrieval-augmented baselines, including \textbf{RAG}, \textbf{Self-RAG}, \textbf{RQ-RAG}, \textbf{RankRAG}, and \textbf{IterDRAG}, under two complementary metrics: (1) the absolute drop in answer accuracy before and after distractor injection, and (2) the distractor citation rate, defined as the proportion of answers that explicitly rely on distractor content during generation.
\begin{figure}[ht]
    \centering
    \hspace*{-0.40cm}
    \includegraphics[scale=0.35]{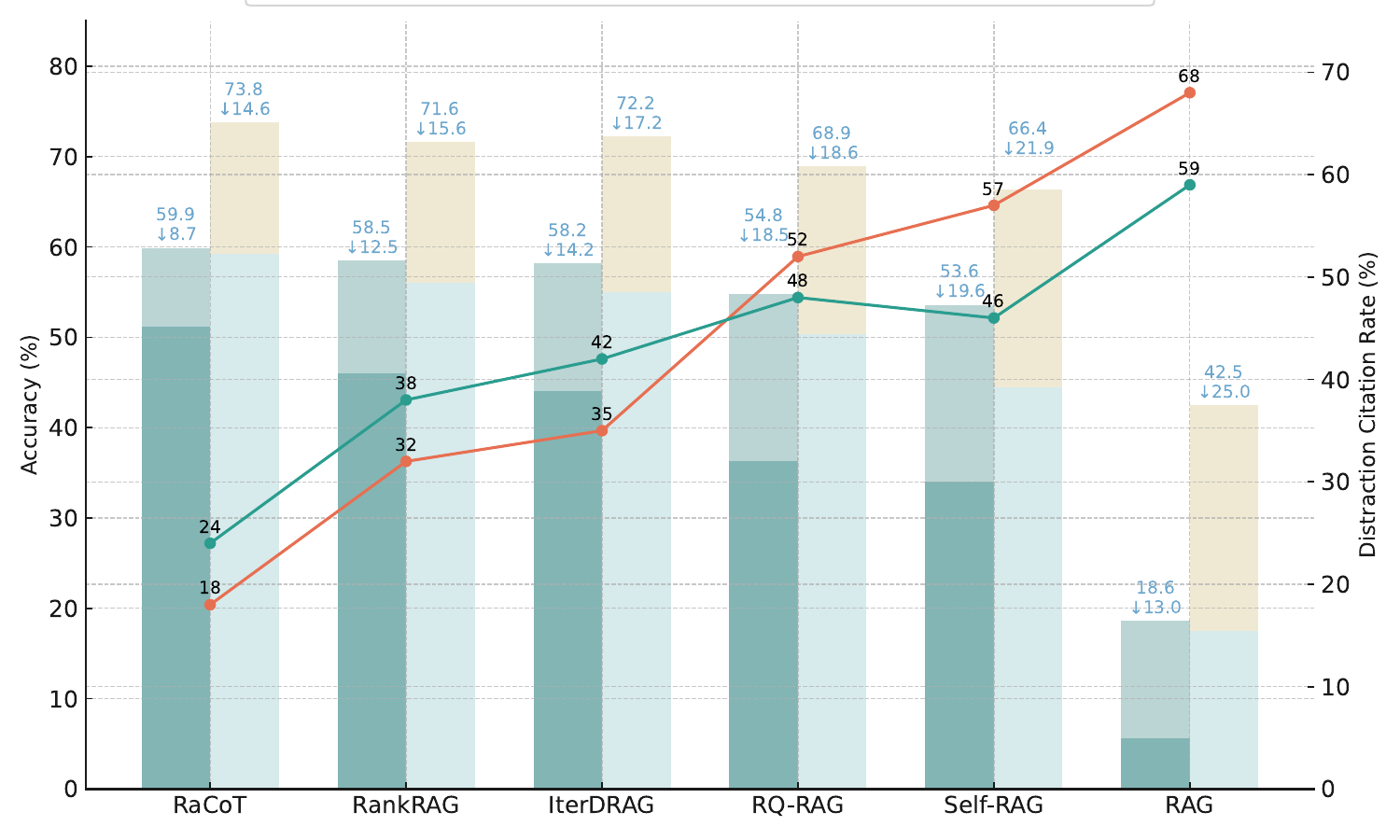} 
    \caption{The result of Retrieval Distractor Confusion shows that the green line indicates accuracy and the orange bars indicate distractor citation rate. } 
    \label{fig:Figure_Retrieval} 

\end{figure}
As shown in Figure~\ref{fig:Figure_Retrieval}, RaCoT consistently achieves the smallest performance drop across both datasets (only $-8.7\%$ on PopQA and $-11.4\%$ on TriviaQA), outperforming the next-best baseline by a large margin. In addition, RaCoT exhibits the lowest distractor citation rate ($18\%$ and $24\%$, respectively), indicating a stronger ability to resist misleading lexical cues. In contrast, other baselines such as Self-RAG and RQ-RAG suffer from substantial degradation (e.g., $-19.6\%$ on PopQA and $-21.9\%$ on TriviaQA) and exhibit high distractor reliance (up to $57\%$). These results validate RaCoT’s core mechanism in mitigating semantic noise and enhancing factual precision under retrieval ambiguity.

\subsubsection{Key Issue Identification Interference}
To further probe the semantic grounding capability of \textbf{RaCoT}, we design a controlled experiment to assess its robustness under semantically perturbed contexts. We randomly sample 5,000 instances each from \textbf{PopQA} and \textbf{ARC-Challenge}, and append contrastive distractor passages to the original queries. These passages are intentionally crafted to be either partially relevant or entirely irrelevant, creating misleading contextual cues that challenge the model's ability to focus on core semantics.

\begin{figure}[ht]
    \centering
    \hspace*{-0.40cm}
    \includegraphics[scale=0.39]{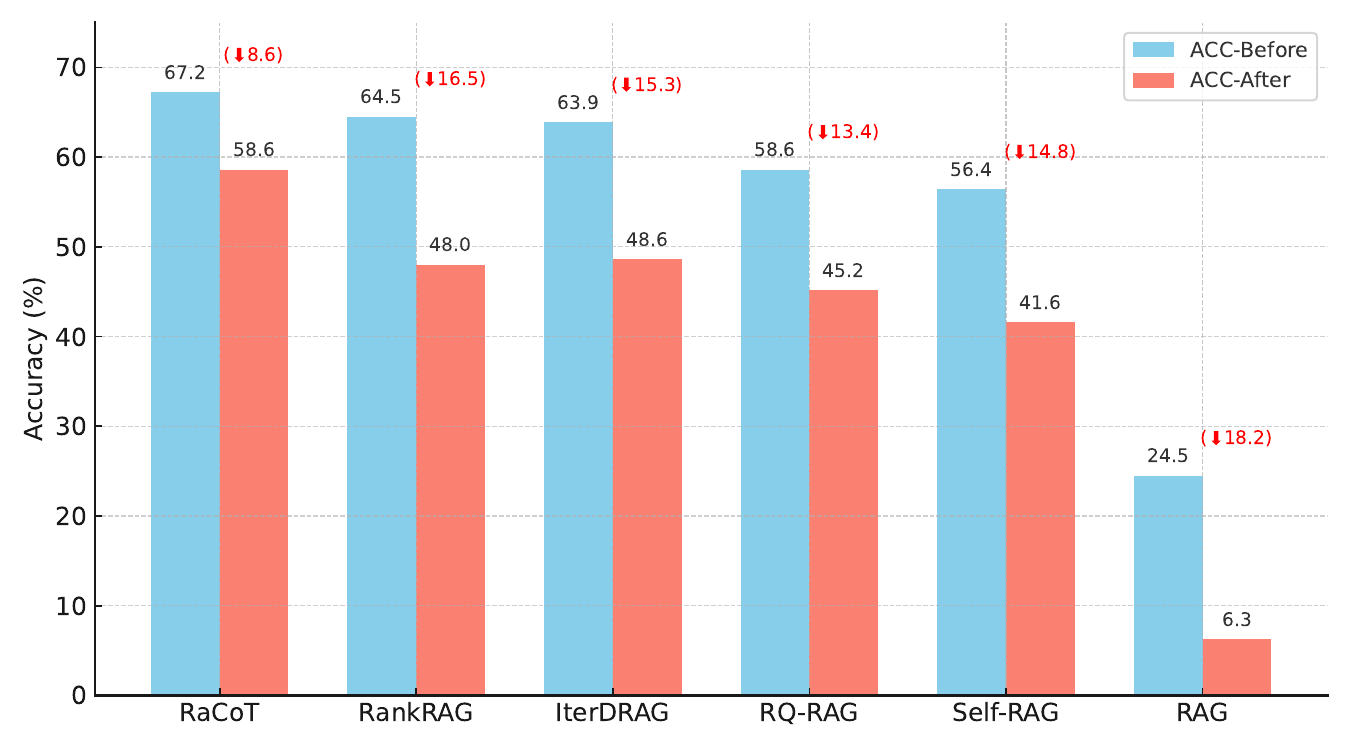} 
    \caption{RaCoT\textsuperscript{+} maintains accuracy under distraction, unlike other RAG methods.} 
    \label{fig:Figure_Key} 

\end{figure}

As shown in Figure~\ref{fig:Figure_Key}, \textbf{RaCoT} achieves a minimal performance degradation of only \textbf{8.6\%}, significantly outperforming all other retrieval-augmented baselines. In comparison, strong competitors such as RankRAG and IterDRAG suffer drops of \textbf{16.5\%} and \textbf{15.3\%}, respectively. Models like Self-RAG and RQ-RAG are even more susceptible, experiencing severe degradation exceeding \textbf{20\%}. This suggests that while existing RAG-based systems can exploit lexical overlap, they often lack the semantic discernment needed to filter out irrelevant content.

The superior robustness of RaCoT highlights its key advantage: by contrasting structurally similar yet semantically distinct passages during retrieval-time prompting, RaCoT effectively anchors reasoning on semantically salient cues and avoids overfitting to surface-level signals. These results reaffirm RaCoT's capability to maintain high factual accuracy even under adversarial or misleading information injection.

\subsection{Efficiency Analysis}
To systematically evaluate the retrieval efficiency and computational overhead of \textbf{RaCoT}, we record both \textit{retrieval latency} and \textit{token consumption} during inference on the \textbf{PopQA} benchmark. For a more comprehensive comparison, we normalize the standard \textbf{RAG} framework to a baseline latency (1.0$\times$), and conduct a controlled evaluation across multiple representative RAG-based methods.
\begin{figure}[ht]
    \centering
    \hspace*{-0.40cm}
    \includegraphics[scale=0.40]{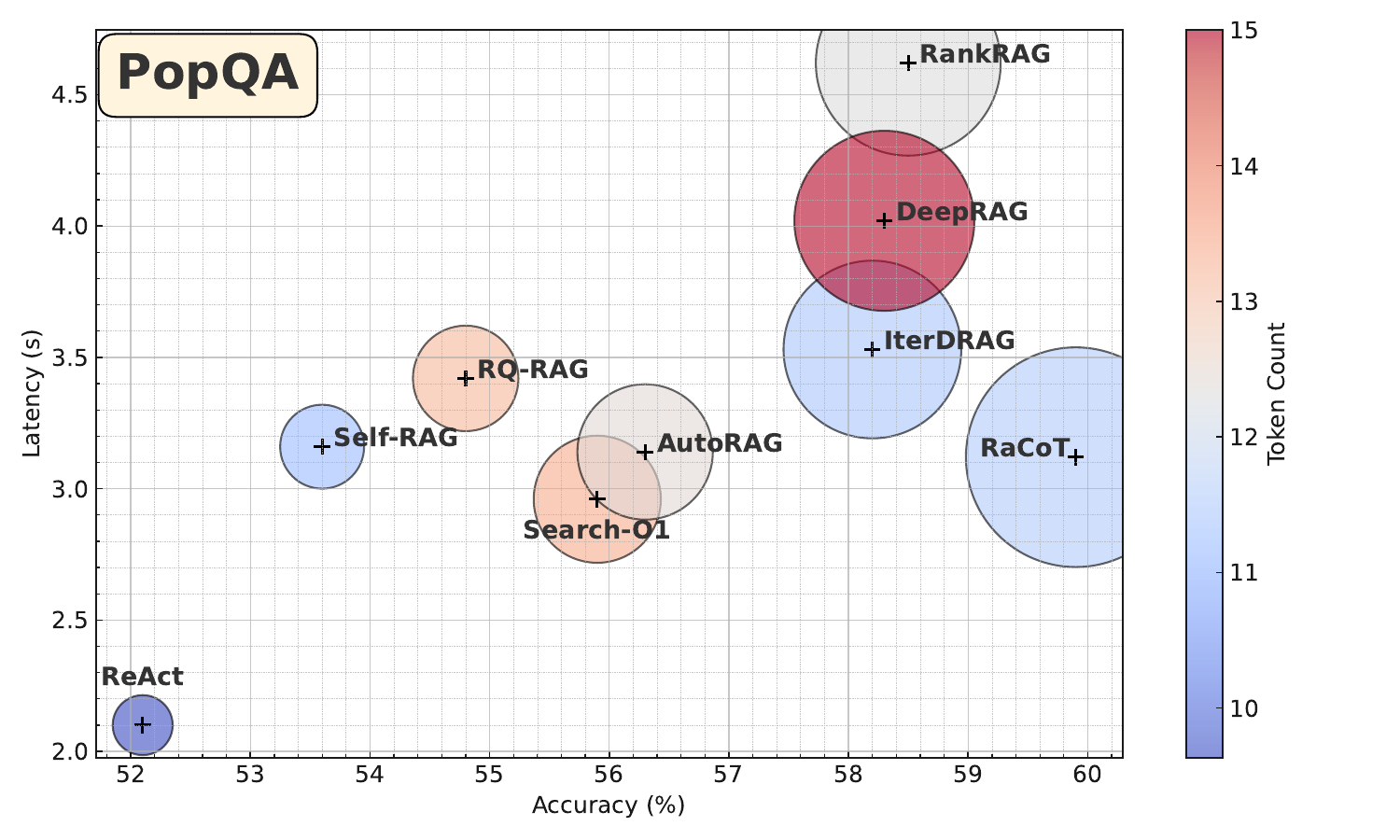} 
    \caption{RaCoT\textsuperscript{+} achieves the highest accuracy with lower latency and token usage. Bubble size denotes answer accuracy, and color reflects average token consumption.} 
    \label{fig:Efficiency Analysis} 

\end{figure}
As shown in Figure~\ref{fig:Efficiency Analysis}, various RAG methods exhibit a clear trade-off among answer accuracy, retrieval latency, and token consumption. Specifically, \textbf{RaCoT} achieves the highest accuracy (59.9\%) while maintaining relatively low latency (3.12 seconds) and the lowest token usage (11.54), demonstrating dual advantages in both effectiveness and efficiency.

In comparison, methods such as \textbf{RankRAG} and \textbf{DeepRAG} attain moderately high accuracy, but suffer from significantly higher latency (4.62s and 4.02s) and increased token costs (12.3 and 15.0) due to their reliance on large-scale candidate ranking and iterative decoding. 
On the other hand, lightweight approaches like \textbf{ReAct} achieve the lowest latency (2.1s) and minimal token consumption (9.63), but perform poorly in terms of accuracy (52.1\%), suggesting limited semantic grounding caused by insufficient retrieval coverage. Methods such as \textbf{IterDRAG} and \textbf{Search-O1} offer a more balanced trade-off between computational cost and answer quality, yet still fall short of the optimal frontier.
Taken together, these results position \textbf{RaCoT} favorably along the accuracy-efficiency Pareto frontier.

\subsection{Ablation Studies}
To evaluate the contribution of each core component in RaCoT, we conduct ablation studies on two long-tail QA benchmarks: PopQA and TQA. The compared variants are as follows:
\textbf{w/o Contrast Prompting}: Removes the $\Delta$-based prompting, using only the original question to evaluate the benefit of explicit contrastive signals.
\textbf{w/o Post-Retrieval Ranking}: Skips the reranking step and directly uses the top-5 retrieved documents to assess the impact of reranking on contextual relevance.
\textbf{w/o Similarity Filtering}: Bypasses the cosine-based filtering during $\Delta$ construction and uses the first teacher-generated question to examine the effect of contrastive sample quality.
\textbf{Weaker Teacher Model}: Replaces the teacher model with a weaker one (e.g., GPT-4o-mini) to assess the model's robustness to teacher strength.

\textbf{Full RaCoT}: The complete system with all components, serving as the performance upper bound.

\begin{figure}[ht]
    \centering
    \includegraphics[scale=0.35]{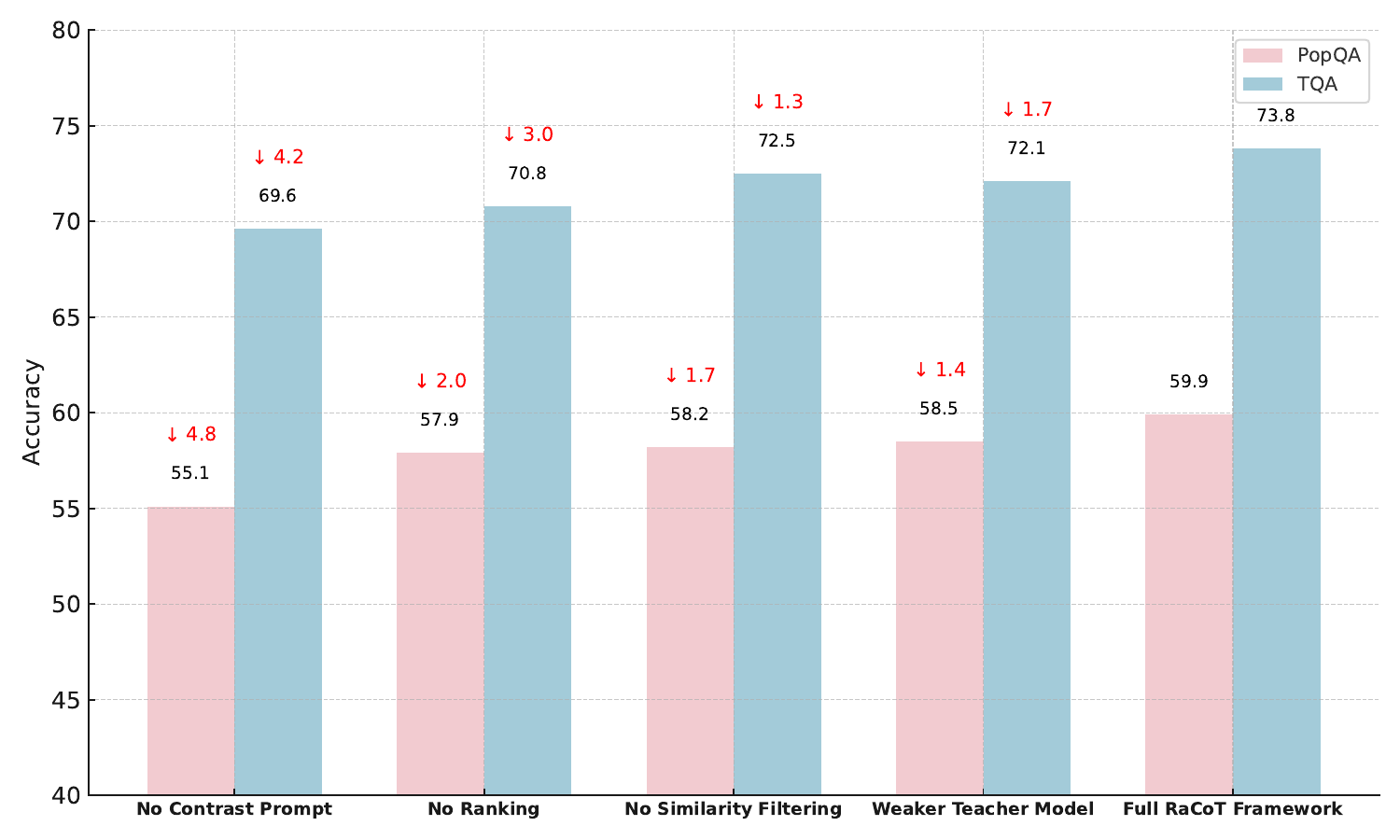} 
    \caption{Results of the ablation studies on PopQA and TQA, demonstrating the performance contribution of each component in RaCoT.} 
    \label{fig:ablation_studies} 
\end{figure}
As shown in Figure~\ref{fig:ablation_studies}, our ablation studies reveal the synergistic contributions of each component. Contrastive prompting is the most critical; its removal causes the largest performance drop (–4.8 on PopQA, –4.2 on TQA), highlighting the value of explicit difference signals. While post-retrieval reranking and similarity filtering offer further gains by refining evidence and ensuring contrast quality, the system remains effective even without them. Moreover, RaCoT's robustness to a weaker teacher model confirms its strength lies in the contrastive mechanism itself, not just teacher capability. Ultimately, the complete RaCoT system achieves the top scores (59.9 on PopQA, 73.8 on TQA), proving its efficacy as a robust pipeline for long-tail QA.

\section{Conclusion}

This paper addresses key limitations in Retrieval-Augmented Generation (RAG) systems, particularly for noisy, ambiguous, and long-tail queries, by reformulating the issue as a semantic bottleneck in query representation. We introduce RaCoT (Retrieval-aware Contrastive-of-Thought), a framework that integrates contrastive reasoning into the retrieval process using semantically adjacent contrastive questions and difference prompts ($\Delta$), enabling active focus on critical evidence in a single pass. Extensive experiments on six QA benchmarks demonstrate RaCoT's superiority over baselines like RankRAG, IterDRAG, and Self-RAG, with gains in accuracy (e.g., +1.5\% on PopQA), robustness to distractors (e.g., 8.6\% minimal drop), and efficiency. Ablation studies confirm the essential roles of contrastive prompting, ranking, and filtering.RaCoT shifts from passive context optimization to robust reasoning restructuring, enhancing practicality for real-world deployments. 

\bibstyle{aaai2026}
\bibliography{aaai2026}

\newpage

\clearpage
\newpage
\appendix
\section{Evaluation Benchmark}

\textbf{PopQA} is a question answering dataset containing a large number of long-tail questions, which involve rare entities and low-frequency knowledge. These questions present a significant challenge for retrieval-based models, as they cover uncommon entities and specialized domains. The sparse knowledge distribution and limited training signals in the dataset require models to possess robust retrieval and understanding capabilities to handle such questions effectively.

\textbf{TriviaQA-unfiltered} is an extended version of the TriviaQA dataset that includes a broader range of unfiltered question-answer data. The dataset spans multiple domains, such as entertainment and history, and features rare entities and low-frequency knowledge. The wide variety and complexity of the questions require retrieval models to handle less common facts and exhibit stronger generalization abilities to effectively address low-frequency issues.

\textbf{ARC-Challenge} is a benchmark dataset designed to evaluate models' reasoning abilities in scientific and commonsense domains. The dataset consists of multiple-choice questions primarily related to subjects like physics and chemistry, aiming to assess whether models can effectively leverage structured scientific knowledge (e.g., formulas and laws) and commonsense knowledge to perform reasoning tasks.

\textbf{OpenBookQA} contains a series of science-related multiple-choice questions that require models to demonstrate not only commonsense reasoning capabilities but also an understanding of open scientific knowledge. This dataset tests how models can leverage scientific knowledge, such as formulas, definitions, and theorems, to solve problems when explicit answers are not provided.

\textbf{HotpotQA} is a multi-hop reasoning dataset that requires models to reason across multiple sentences or documents. The questions often involve multiple information sources, and models must extract and integrate information from them while tracking complex reasoning chains. HotpotQA is designed to assess models' ability to handle and reason across document-level information.

\textbf{2WikiMultiHopQA} is a question answering dataset that focuses on multi-hop reasoning across multiple Wikipedia documents. It tests the ability of models to perform reasoning across different documents, requiring them to find relevant data from multiple information sources and integrate it effectively. This dataset assesses models' ability to understand complex reasoning chains and integrate information from multiple documents.

\section{Model Configuration and Implementation Details}

The RaCoT framework is built upon the classical Retrieval-Augmented Generation (RAG) paradigm and consists of four core stages: contrastive example generation, semantic difference extraction, enhanced retrieval representation construction, and contrast-aware generation. We adopt an open-source, end-to-end implementation aligned with mainstream RAG architectures to facilitate fair and reproducible comparisons.

\textbf{Retriever.} We utilize ColBERTv2, a domain-adapted dense retriever that supports fine-grained matching and exhibits strong alignment capabilities on long-form documents. To remain compatible with standard RAG baselines, we additionally incorporate BM25 as a sparse retriever. The two retrieval results are fused using a dual-fusion mechanism, where the top-$K$ passages from both retrievers are merged and deduplicated ($K=20$). The retrieval corpus consists of either the English Wikipedia dump (2023 edition) or the ``Chinese Legal Knowledge Graph v1.3,'' segmented into non-overlapping 100-token chunks.

\textbf{Encoder.} We employ a standard dual-encoder architecture based on HuggingFace Transformers to compute semantic embeddings for queries and documents, ensuring vector alignment for retrieval. To support the contrastive reasoning required by RaCoT, we enhance the original query $Q$ by incorporating a contrastive variant $Q'$ and its semantic difference description $\Delta$, resulting in an enriched retrieval representation $Q^*$. In practice, the triplet is concatenated as a natural language prompt: ``Original question: \{...\}; Contrastive question: \{...\}; Semantic difference: \{...\}'', and encoded using the existing transformer-based encoder without modification.

\textbf{Generator.} For generation, we employ two representative large language models: \textbf{LLaMA2-13B-chat} and \textbf{Qwen2.5-7B}. These models are used independently in separate experiments to verify RaCoT’s model-agnostic compatibility. We follow the standard retrieve-then-generate workflow. Prior to generation, candidate documents are scored using RaCoT’s lightweight relevance ranking module, and only those with scores above a threshold ($\tau = 0.7$) are retained as the final context.

All modules are implemented using PyTorch and HuggingFace Transformers. Dense retrieval is accelerated using \textbf{Faiss}, with GPU-based batched inference enabled. All experiments are conducted on 4$\times$NVIDIA A100 GPUs with FP16 mixed-precision training to optimize memory usage and throughput.

\section{Retrieval Quality}

To evaluate retrieval quality under a limited document budget ($K = 10$), we compare four retrieval strategies: RAG (standard single-pass retrieval), RankRAG (post-retrieval reranking), IterDRAG (multi-step query decomposition), and RaCoT (contrast-enhanced retrieval). Experiments are conducted on three QA benchmarks: \textbf{HotpotQA} (with gold supporting evidence), \textbf{TriviaQA-unfiltered}, and \textbf{PopQA}, each with 100 randomly sampled test questions.

All methods use a shared retrieval corpus (Wikipedia 2023) and a hybrid retriever combining BM25 and ColBERTv2. RAG, RankRAG, and RaCoT retrieve top-10 documents in a single pass; IterDRAG performs up to three rounds of subquery-based retrieval, retrieving 5 documents per step and merging the results.

We evaluate the top-10 retrieved documents from four perspectives:

\begin{itemize}
    \item \textbf{AutoRel@10 (automatic relevance).} We use ChatGPT-4o to rate each retrieved document on a 1--5 scale based on semantic relevance to the question. Scores are averaged across the top-10 results.
    
    \item \textbf{HumanRel@10 (human factual relevance).} Three expert annotators rate whether each document contains factual content necessary to answer the question. Scores are on a 1--5 Likert scale.
    
    \item \textbf{HumanUse@10 (human utility).} Annotators also assess the practical usefulness of each document for human problem-solving, again using a 1--5 scale.
    
    \item \textbf{Recall@10.} For HotpotQA, we report whether any gold-supporting evidence document appears in the top-10 set.
\end{itemize}

Human evaluations are conducted in a blind setting with unified scoring criteria and calibration examples. Finally, we compute a composite score, \textbf{RetrievalUtility@10}, defined as a weighted sum of AutoRel@10 and HumanUse@10 (default weights: 0.6 and 0.4), reflecting both semantic alignment and practical downstream value.

\begin{table}[H]
\caption{Retrieval quality comparison across QA datasets (Top-10 documents). RaCoT outperforms retrieval-enhanced baselines in both automatic and human evaluations.}
\label{tab:retrieval_quality_racot}
\centering
\renewcommand{\arraystretch}{1.1}
\footnotesize
\resizebox{\columnwidth}{!}{%
\begin{tabular}{lcccccc}
\toprule
\textbf{Method} & \textbf{AutoRel@10} & \textbf{HumanRel@10} & \textbf{HumanUse@10} & \textbf{Recall@10} & \textbf{Utility@10} \\
\midrule
\multicolumn{6}{c}{\textit{HotpotQA}} \\
RAG & 3.48 & 3.20 & 3.01 & 0.48 & 3.20 \\
RankRAG & 3.61 & 3.36 & 3.18 & 0.55 & 3.34 \\
IterDRAG & 3.74 & 3.50 & 3.32 & 0.63 & 3.49 \\
\rowcolor[HTML]{D9D9D9} \textbf{RaCoT (Ours)} & \textbf{3.90} & \textbf{3.65} & \textbf{3.47} & \textbf{0.66} & \textbf{3.73} \\
\midrule
\multicolumn{6}{c}{\textit{TriviaQA-unfiltered}} \\
RAG & 3.00 & 2.84 & 2.69 & -- & 2.81 \\
RankRAG & 3.20 & 3.01 & 2.88 & -- & 3.07 \\
IterDRAG & 3.38 & 3.18 & 3.02 & -- & 3.16 \\
\rowcolor[HTML]{D9D9D9} \textbf{RaCoT (Ours)} & \textbf{3.52} & \textbf{3.28} & \textbf{3.15} & -- & \textbf{3.27} \\
\midrule
\multicolumn{6}{c}{\textit{PopQA}} \\
RAG & 3.12 & 2.94 & 2.78 & -- & 2.92 \\
RankRAG & 3.29 & 3.10 & 2.89 & -- & 3.05 \\
IterDRAG & 3.46 & 3.25 & 3.07 & -- & 3.22 \\
\rowcolor[HTML]{D9D9D9} \textbf{RaCoT (Ours)} & \textbf{3.60} & \textbf{3.38} & \textbf{3.21} & -- & \textbf{3.45} \\
\bottomrule
\end{tabular}%
}
\end{table}

As shown in Table~\ref{tab:retrieval_quality_racot}, RaCoT achieves the best overall performance across all datasets and evaluation metrics. On HotpotQA, it reaches a Recall@10 of 66\%, representing an 18\% improvement over vanilla RAG, indicating significantly stronger grounding capability for retrieving gold supporting evidence. Additionally, RaCoT consistently outperforms all baselines on AutoRel, HumanRel, and HumanUse, suggesting that its contrast-enhanced query formulation aligns more closely with the question intent and facilitates the retrieval of more useful content. While IterDRAG performs competitively on some tasks, its reliance on multi-step decomposition introduces added complexity and latency. In contrast, RaCoT achieves superior results with a single-pass retrieval, offering a favorable trade-off between quality and efficiency. Overall, RaCoT demonstrates robust retrieval utility even under a strict Top-10 document constraint, providing stronger evidence grounding and higher-quality input for downstream QA generation.

\section{Attention Focus Analysis}
To assess the focusability of different retrieval methods during generation, we conduct a cross-attention–based evaluation. We use LLaMA2-13B-chat as the generator and feed in the top-10 retrieved documents from each method: RAG, RankRAG, IterDRAG, and RaCoT. The analysis is performed on 100 randomly sampled questions from the HotpotQA test set. For each question, gold supporting sentences are extracted from existing annotations.

During generation, we log the cross-attention weights from the decoder layers and align them with the gold evidence tokens to evaluate whether the model attends to relevant content. All generations are produced using greedy decoding with identical temperature and prompt formatting to ensure fairness across methods.

We adopt three widely used metrics for attention analysis: \textbf{Attention Over Evidence (AOE)}, \textbf{Top-K Hit Rate@30}, and \textbf{Average Attention Entropy}. These metrics provide complementary views on how concentrated the attention is on relevant context tokens, thereby reflecting the quality and semantic steerability of the retrieved documents.

\begin{table}[H]
\caption{Attention focus comparison on HotpotQA using decoder cross-attention. RaCoT achieves stronger evidence alignment across all official attention metrics.}
\label{tab:attention_focus_racot}
\centering
\renewcommand{\arraystretch}{1.1}
\footnotesize
\resizebox{\columnwidth}{!}{%
\begin{tabular}{lccc}
\toprule
\textbf{Method} & \textbf{Attention Over Evidence (↑)} & \textbf{TopK-Hit@30 (↑)} & \textbf{Attention Entropy (↓)} \\
\midrule
RAG         & 0.218 & 49.2\% & 3.97 \\
RankRAG     & 0.243 & 54.6\% & 3.85 \\
IterDRAG    & 0.264 & 57.9\% & 3.71 \\
\rowcolor[HTML]{D9D9D9} \textbf{RaCoT (Ours)} & \textbf{0.291} & \textbf{61.3\%} & \textbf{3.56} \\
\bottomrule
\end{tabular}%
}
\end{table}

As shown in Table~\ref{tab:attention_focus_racot}, RaCoT achieves the best performance across all three attention metrics. Its Attention Over Evidence (AOE) reaches 0.291, noticeably higher than IterDRAG (0.264) and RankRAG (0.243), indicating stronger alignment with gold supporting content during generation. On the TopK-Hit@30 metric, RaCoT attains a hit rate of 61.3\%, outperforming RAG by over 12 percentage points, which suggests that the retrieved context is more semantically salient and easier for the decoder to recognize. Additionally, RaCoT achieves the lowest attention entropy (3.56), reflecting a more focused distribution over relevant tokens. These results demonstrate that RaCoT effectively improves semantic grounding in generation by guiding the model to attend to higher-quality evidence.

\begin{figure*}[t] 

\begin{tcolorbox}[
    title=Prompt: Contrastive Question Generation,
    colback=gray!5!white,
    colframe=gray!75!black,
    coltitle=white,
    fonttitle=\bfseries,
    arc=2mm,
    boxrule=0.8pt
]
\small
You are an expert AI assistant specialized in creating high-quality, challenging test cases.

\vspace{0.5em}
\textbf{1. Context and Goal:}
Your mission is to generate a "semantic trap" for a given \textit{Original Question}. This involves creating a \textit{Contrastive Question} that is deceptively similar in topic and wording, but whose answer is fundamentally different due to a single, critical change in meaning. This process helps to rigorously test an AI's attention to semantic detail.

\vspace{0.5em}
\textbf{2. Your Task:}
You must produce two outputs based on the original question provided:
\begin{enumerate}[leftmargin=1.5em]
    \item \textbf{A Contrastive Question:} A new question that is semantically close to the original but has a verifiably different answer.
    \item \textbf{A Key Difference:} A concise label that pinpoints the exact semantic element causing the divergence in answers.
\end{enumerate}

\vspace{0.5em}
\textbf{3. Guiding Principles:}
\begin{itemize}[leftmargin=1.5em]
    \item \textbf{High Similarity:} The contrastive question must be a plausible variation that reuses the original's vocabulary and structure.
    \item \textbf{Single, Critical Difference:} The change must be isolated to one element (e.g., a role, an attribute, a number, a concept).
    \item \textbf{Verifiably Different Answers:} The two questions must have objectively different and correct answers.
    \item \textbf{Concise Difference Label:} The key difference must strictly follow the format \texttt{"\{Element A\} vs. \{Element B\}"}.
\end{itemize}

\vspace{0.5em}
\textbf{Required Output Format:}
Your response must be a single, raw JSON object adhering to the following structure. \textbf{Do not} include any explanations, comments, or markdown formatting (like \texttt{```json}). Just the JSON object itself.
\begin{verbatim}
{
  "original_question": "The original question text here.",
  "contrastive_question": "Your generated contrastive question here.",
  "key_difference": "Your key difference label here."
}
\end{verbatim}

\vspace{0.5em}
\textbf{Examples:}

\vspace{0.3em}
\textbf{Example 1: Conceptual Distinction} \\
\textit{Original Question: Why did Renaissance paintings emphasize the representation of space and reality more than those from the Middle Ages?}
\begin{verbatim}
{
  "contrastive_question": "Why did Renaissance paintings emphasize realism more than those 
  from the Middle Ages?",
  "key_difference": "{representation of space and reality} vs. {realism}"
}
\end{verbatim}

\vspace{0.3em}
\textbf{Example 2: Role Distinction} \\
\textit{Original Question: Who directed the movie Inception?}
\begin{verbatim}
{
  "contrastive_question": "Who starred in the movie Inception?",
  "key_difference": "{directed} vs. {starred in}"
}
\end{verbatim}

\vspace{0.3em}
\textbf{Example 3: Attribute Distinction} \\
\textit{Original Question: What is the capital city of Australia?}
\begin{verbatim}
{
  "contrastive_question": "What is the largest city in Australia?",
  "key_difference": "{capital city} vs. {largest city}"
}
\end{verbatim}

\vspace{0.3em}
\textbf{Example 4: Ordinal Distinction} \\
\textit{Original Question: Who was the second US astronaut to walk on the moon?}
\begin{verbatim}
{
  "contrastive_question": "Who was the first US astronaut to walk on the moon?",
  "key_difference": "{second} vs. {first}"
}
\end{verbatim}

\vspace{0.5em}
\textbf{Now, generate the contrastive pair for the question provided below.}

\end{tcolorbox}

\end{figure*}

\begin{figure*}[t]

\begin{tcolorbox}[
    title=Question Answering Prompt,
    colback=gray!5!white,
    colframe=gray!75!black,
    coltitle=white,
    fonttitle=\bfseries,
    arc=2mm,
    boxrule=0.8pt
]
\small

\noindent\textbf{System:} This is a chat between a user and an artificial intelligence assistant. The assistant provides helpful, detailed, and polite answers based on the context. The assistant should also indicate when the answer cannot be found in the context.

\vspace{1.5em}

\noindent\textbf{CONTEXT PASSAGES:}
\vspace{0.5em}
\begin{quote}
    \textbf{Passage 1:} \textit{\{Retrieved Passage 1\}} \\
    \textbf{Passage 2:} \textit{\{Retrieved Passage 2\}} \\
    \textbf{Passage 3:} \textit{\{Retrieved Passage 3\}} \\
    \textbf{...}
\end{quote}

\vspace{1em}

\noindent\textbf{User:} \textit{\{Question\}}

\vspace{1em}

\noindent\textbf{Assistant:} Based on the retrieved relevant passages, I will provide an answer. If there is any inconsistency or if no clear answer can be found, I will indicate it.

\noindent\textbf{Answer the following question based on the passages above: \{Question\}}

\end{tcolorbox}

\end{figure*}
\end{document}